\documentclass[11pt]{article}

\usepackage{acl}
\usepackage{times}
\usepackage{latexsym}
\usepackage[T1]{fontenc}
\usepackage[utf8]{inputenc}
\usepackage{microtype}
\usepackage{inconsolata}
\usepackage{graphicx}
\usepackage{booktabs}
\usepackage{enumitem}
\usepackage{tabularx}
\usepackage{array}

\title{Project \textsc{Kaleidoscope}: Contextual, Human-Aligned Evaluation for Real-World AI Applications}

\author{
 \textbf{Leanne Tan\textsuperscript{1}\thanks{Contributed equally}},
 \textbf{Rohan Jaggi\textsuperscript{2}\thanks{Work done during an internship at GovTech Singapore.}},
 \textbf{Shaun Khoo\textsuperscript{1}},
 \textbf{Roy Ka-Wei Lee\textsuperscript{1,3}},
\\
\\
    \textsuperscript{1}GovTech, Singapore,
    \textsuperscript{2}National University of Singapore,
    \textsuperscript{3}University of British Columbia \\
    \texttt{leanne\_tan@tech.gov.sg, roy.lee@ubc.ca}\\
}

\begin{document}
\maketitle
\begin{abstract}
Evaluations (Evals) are a deployment bottleneck for real-world AI applications: public benchmarks rarely match a team's users, context, or policies, and human review is often tedious to scale. Motivated by our work with AI applications in the public sector, this project addresses recurring evaluation challenges encountered when applications must satisfy local policy and governance requirements. We present \textsc{Kaleidoscope}, an integrated workflow for contextual functional evaluation that links persona-based test generation, contextualized rubrics, and human review for reliability-gated automated scoring. Generated test cases are scored against application-specific rubrics; human annotations provide reviewable labels; and LLM judges automate scoring only when their agreement with those labels meets a configured threshold. \textsc{Kaleidoscope} is therefore a practical, inspectable, iterative workflow for product teams. We report early evidence from a three-week pilot across four organizational use cases and custom-rubric judge experiments on 108 annotated Q\&A pairs spanning four domains and 14 evaluation dimensions. The results highlight useful features for end-to-end reliable, automated scoring.

\end{abstract}

\section{Introduction}

Large language model (LLM) applications are increasingly easy to prototype, but harder to evaluate in the setting in which they will be deployed. In this paper, we focus on \emph{functional evaluation}: task-specific assessment of whether an AI application response satisfies the user and policy requirements for a given use case. This is distinct from general capability benchmarking and broader safety or security assessment, although these concerns may constrain the rubric for a given deployment. Functional evaluation is especially difficult because what counts as ``good'' depends on the application context, users, policies, workflow, and risk tolerance. For example, a math learning chatbot may need step-by-step worked solutions, while an HR policy assistant may need concise, policy-grounded answers.

In organizational contexts, application owners rely on manual, fragmented evaluation processes in which teams curate tests, run them through their applications, and then slowly review long responses. More informed engineers may reuse general evaluation benchmarks or generate synthetic tests, but these can be repetitive, unrealistic, or misaligned with application-specific behaviors that matter in deployment \citep{openai2025evals}. Existing tools, benchmark suites, and LLM-as-judge methods address parts of this workflow, but the remaining gap is the integration of contextual test construction, configurable rubrics, human calibration, reliability-gated LLM judging, and error analysis into a single inspectable workflow.

These operational constraints motivated \textsc{Kaleidoscope}, an evaluation workflow for real-world AI applications. Rather than introducing a new benchmark or a universal evaluation model, \textsc{Kaleidoscope} is designed around the observation that contextual evaluation requires careful test construction, domain-specific quality criteria, and automated judging calibrated against human-reviewed labels before scores are aggregated. These reliability checks help guide automated scoring decisions.

We make three contributions: (i) We present \textsc{Kaleidoscope}\footnote{Code repositor and documentation can be accessed via: \href{https://govtech-responsibleai.github.io/kaleidoscope/}{govtech-responsibleai.github.io/kaleidoscope}}, an end-to-end workflow artifact for contextual functional evaluation of AI applications, covering target setup, persona-based test generation, application output collection, human review, automated judging, and error analysis. (ii) We describe its rubric and judging design, including configurable evaluation criteria, claim- or response-level scoring, and reliability-gated aggregation of LLM judge scores using human-reviewed calibration labels. (iii) We report early empirical lessons and limitations of a three-week pilot in four organizational use cases and rubric/judge experiments on 108 annotated Q\&A pairs spanning four domains and 14 evaluation dimensions.

\section{Related Work}

\paragraph{Benchmarks and contextual evaluation metrics.}
General-purpose benchmarks help compare model capabilities, but in practice, they rarely capture the specific users, policies, workflows, and risk tolerance of a deployed AI application. Prior work has moved from standard coding, language, reasoning, or cybersecurity tasks to behavioral and task-oriented criteria. For instance, HELM proposes the use of holistic, multi-dimensional evaluation across risks and use contexts \citep{liang2023holisticevaluationlanguagemodels} while other work measures \textit{sensibleness} and \textit{specificity} in open-domain dialogue, \textit{answerability} for chatbot responses, and \textit{empathy} in conversational systems \citep{adiwardana2020meena,gupta2022answerability,chen2024emotionqueen,krishna2023longeval}. Beyond these specialized metrics, recent contextualized evaluation work further shows that under-specified queries can force arbitrary judgments and produce inconsistent conclusions about model quality \citep{ai2contextualized2025}. These findings motivate \textsc{Kaleidoscope}'s emphasis on custom, contextual criteria. Unlike benchmarks or leaderboards, \textsc{Kaleidoscope} produces application-specific evaluations whose scores support local governance decisions, not direct comparison across applications.

\paragraph{LLM-as-a-judge and Human Review.}
LLM evaluators can scale scoring against natural-language criteria; G-Eval shows that structured prompts can help large models approximate human judgment for generation tasks \citep{liu2023geval}. Yet LLM judges can show position, style, length, or concreteness biases \citep{gu2025surveyJudge}, low intra-rater reliability even under nominally deterministic settings \citep{haldar2025rating}, and agreeable bias with high true-positive but low true-negative rates \citep{jain2025beyond}. Agreement with human judgment, measured through metrics such as percentage agreement, Cohen's kappa, and rank correlation, therefore remains a common reliability check \citep{gu2025surveyJudge}; however, human-in-the-loop (HITL) review is costly and can suffer fatigue-related inconsistencies \citep{pignatiello2020decisionFatigue,karim2025annotationAgents}. Related work on building human expert datasets suggests that audit-oriented review processes can lead to more reliable annotations. \citep{saligrama2026groundTruthProcess}. Together, these findings motivate a focus on human reviewability and on reducing annotation fatigue when calibrating automated evaluations. A complementary practitioner pattern is LLM-as-a-jury, where multiple evaluator models expose judge variation and disagreement \citep{arizeJury}. These considerations inform \textsc{Kaleidoscope}'s reliability-gated jury design in Section~\ref{subsec:llm-jury}.

\paragraph{Evaluation tools.}
Existing tools support many parts of the eval lifecycle, including defining, running, scoring, and logging evals. OpenAI Evals and Inspect AI provide reusable evaluation workflows for LLM systems \citep{openaiEvalsGithub,inspectAI}, Promptfoo supports test cases, assertions, and regression checks \citep{promptfoo}, and EvalAssist helps users create and refine LLM-as-a-judge evaluations \citep{ibmEvalAssist}. In our organizational context, however, teams still needed a way to construct context-specific test sets, translate domain goals into customizable criteria, and make automated scoring transparent, inspectable, and gated on judge alignment with human-reviewed labels. \textsc{Kaleidoscope} therefore assembles ideas from existing evaluation research, user feedback, and organizational context into an end-to-end workflow for contextual functional evaluation.

\section{Design Considerations}

The design requirements for \textsc{Kaleidoscope} follow from the constraints described above: product teams need evaluation artifacts that reflect their target application, can be reviewed and calibrated by humans, and can expose when automated scoring is or is not appropriate. Rather than proposing a universal benchmark or a general solution to LLM-judge reliability, \textsc{Kaleidoscope} supports contextual functional evaluation of a specific AI application. Its scores are intended to be interpreted locally, with respect to the application under test, the configured rubric, and the calibration set used to assess judge alignment.

Accordingly, \textsc{Kaleidoscope} operationalizes these requirements through four design considerations.

\paragraph{R1: Representative test cases.}
Evaluation data should approximate realistic users, tasks, and contexts for the target AI system.

\paragraph{R2: Configurable evaluation rubrics.}
Different agencies have different policy requirements. The workflow should translate evaluation goals into specific, explicit, reviewable scoring criteria.

\paragraph{R3: HITL User Experience.}
If human review is required for calibration and high-risk applications, an evals workflow should optimize for human reviewability while minimizing reviewer fatigue.

\paragraph{R4: Transparent, reliability-gated automated scoring.}
LLM judge behavior should be transparent and reliable because automated scoring requires governance confidence. The workflow should aggregate automated scores only when a judge meets a certain alignment threshold on a human-reviewed subset, and should support multiple judges to surface disagreement and error analysis.

\section{The \textsc{Kaleidoscope} Workflow}

\begin{figure}[t]
  \centering
  \includegraphics[width=0.98\columnwidth]{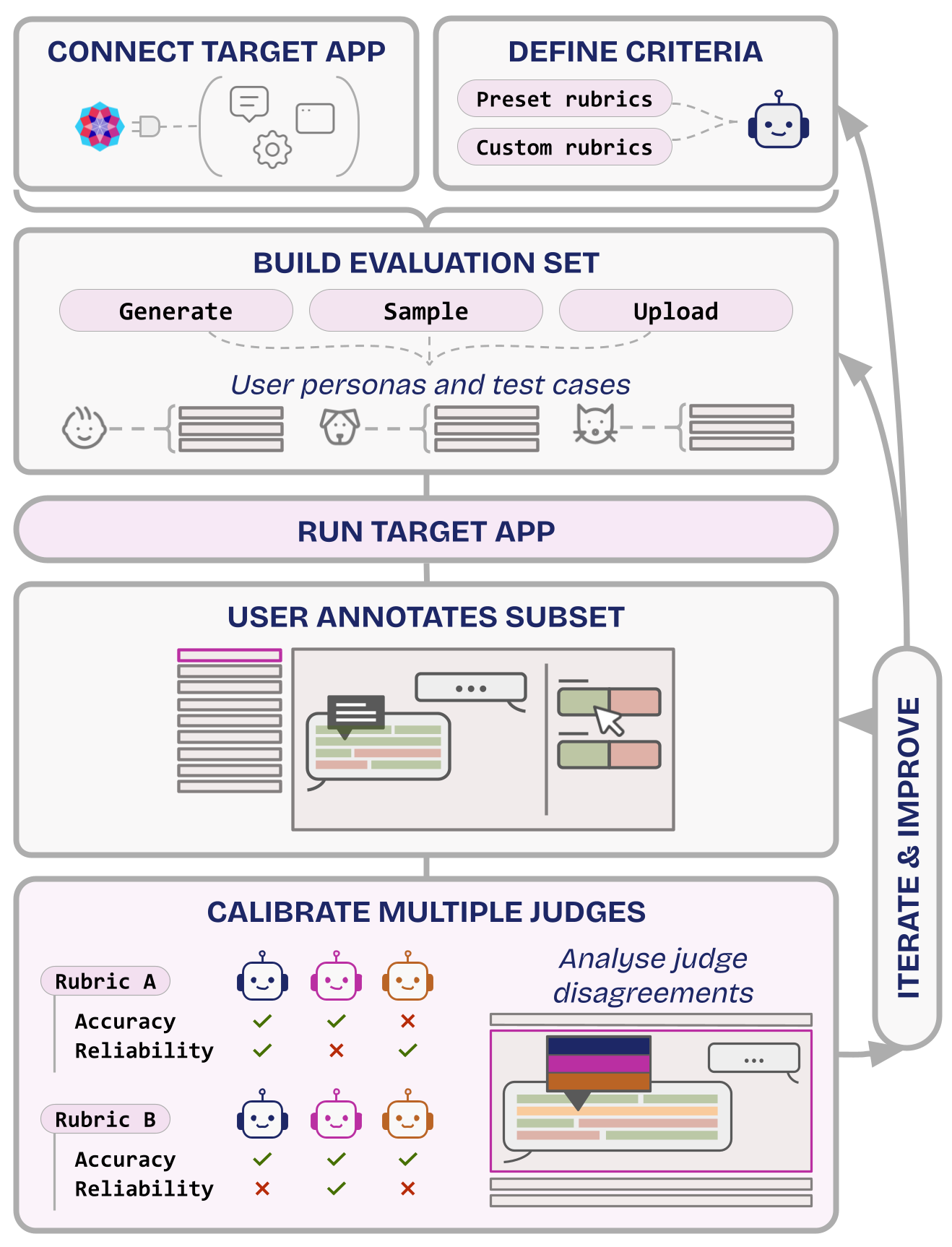}
  \caption{\textsc{Kaleidoscope}'s workflow.}
  \label{fig:workflow}
\end{figure}

\textsc{Kaleidoscope} implements these requirements as an end-to-end workflow (Figure~\ref{fig:workflow}). The workflow separates three artifacts: the evaluation set, the human labels used for calibration and quality control, and automated scores produced after calibration. 

\subsection{Defining the Target Application}

The user specifies the application under evaluation, including its organization, purpose, target users, and knowledge base. This application profile grounds three downstream steps:  (i) generating domain-appropriate test cases,  (ii) executing test cases against the target application and collecting responses, and (iii) supporting preset rubrics such as groundedness checks against source material.

When user-provided context is sparse, the backend can optionally retrieve public web information about the application to give the generator additional grounding. Users also define evaluation rubrics and goals during setup, as described in Section~\ref{subsec:evaluation-rubrics}. At runtime, \textsc{Kaleidoscope} instantiates candidate LLM judges from these criteria and evaluates them against the human-labeled calibration set before aggregation.

\subsection{Persona-Based Test Generation}

Users may upload an evaluation set or utilize \textsc{Kaleidoscope} to generate one. Generation varies questions along the five configurable dimensions specified in Table~\ref{tab:generation-dimensions} to improve domain coverage and user variation.

\begin{table}[h]
    \centering
    \scriptsize
    \setlength{\tabcolsep}{3pt}
        \begin{tabular}{@{}p{0.15\columnwidth}p{0.33\columnwidth}p{0.46\columnwidth}@{}}
        \toprule
        \textbf{Dim} & \textbf{Values} & \textbf{Goal} \\
        \midrule
        Persona & Selected / Defined by User & Test cases are distributed across selected user personas. \\
        Type & Typical / Edge & Covers common and boundary-testing inputs.  \\
        Scope & In / Out KB & In-KB inputs use uploaded documents; out-KB inputs test unsupported requests. \\
        Input style & Brief / Regular / Detailed & Varies input formality and length. \\
        Language & Selected / Defined by user & Splits questions across requested languages. \\
        \bottomrule
    \end{tabular}
    \caption{Dimensions diversifying generated test cases.}
    \label{tab:generation-dimensions}
\end{table}

Personas provide the user model for this diversification. Each persona specifies background, communication style, and product use case. Personas can be generated from the target-application profile, sampled from open-source persona datasets~\citep{nvidiaNemotronPersonas}, or entered manually. After configuration, the system allocates questions across the selected dimensions using configurable default ratios described in Appendix~\ref{app:test-generation}.

\subsection{Human Review and Annotation}
\label{subsec:human-review}

\textsc{Kaleidoscope} collects all responses from the target application and guides reviewers to label a review set for quality control. This set may contain all responses in a small evaluation or a subset of responses in a larger evaluation. The rubric determines the annotation unit: claim-level rubrics are labeled on individual extracted claims, while response-level rubrics receive one label for the response as a whole. Additional claim-level implementation details are provided in Appendix~\ref{app:claim-level-scoring}.

\begin{figure}[t]
  \centering
  \includegraphics[width=0.8\columnwidth]{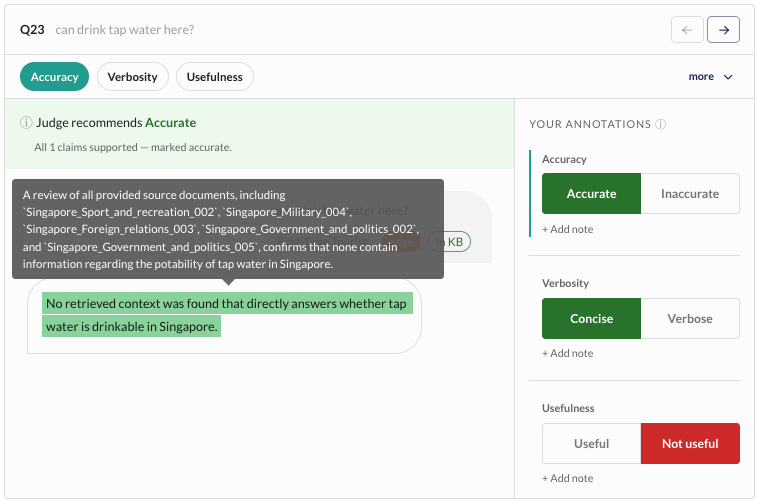}
  \caption{LLM-assisted review interface showing highlighted evidence and judge verdicts.}
  \label{fig:claim-highlighter}
\end{figure}

The review interface incorporates an LLM-assisted annotation workflow, displaying candidate labels and highlighted claims from a first-pass LLM judge (Figure~\ref{fig:claim-highlighter}). This shifts the reviewer task from one-shot labeling towards auditing suggested annotations to improve annotation reliability. To reduce automation bias, reviewers must select a label from the rubric options instead of simply approving or rejecting a judge label. 

Human review is used to establish a trusted reference set for downstream judge alignment with human judgment. Only judges that meet the configured threshold are included in the aggregated automated score described in Section~\ref{subsec:llm-jury}.

\subsection{Evaluation Rubrics}
\label{subsec:evaluation-rubrics}
During setup, users either select preset rubrics for common organizational evaluation needs or define custom rubrics with natural-language criteria, scoring options, and annotation level. 

This design was informed by experiments on a hand-labeled set spanning four organizational use cases and 14 evaluation criteria (Appendix~\ref{app:rubric-experiments}). In brief, generic rubric insertion was not reliable enough across metrics, and asking one judge prompt to score multiple rubric dimensions substantially degraded performance. Accordingly, \textsc{Kaleidoscope} uses presets where available, a one-metric-per-judge-prompt design for custom rubrics, and an LLM-based augmenter that rewrites user-defined criteria into more structured judge prompts before scoring.

\subsection{Automated Scoring and LLM-as-a-Jury}
\label{subsec:llm-jury}

For each rubric, \textsc{Kaleidoscope} creates three candidate judges using the user's configured model providers. Each judge scores the relevant annotation units, and its alignment is measured only on the human-labeled calibration units. 

\begin{figure}[t]
  \centering
  \includegraphics[width=0.98\columnwidth]{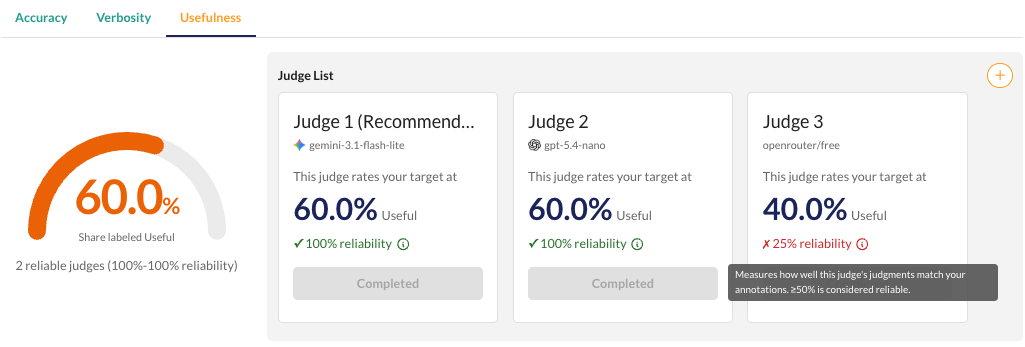}
  \caption{Screenshot of 3 LLM Judges. Only reliable judges are aggregated.}
  \label{fig:llm-jury}
\end{figure}

Only judges that pass the local reliability gate of Macro F1 > 0.5 are eligible for majority-vote aggregation shown in Figure~\ref{fig:llm-jury}. If no judge passes, the system withholds an automated aggregate for that rubric and flags it for further human review or rubric/judge reconfiguration. The gate is not meant to be evidence of universal judge correctness, but a local quality-control check against human labels.

\begin{figure}[t]
  \centering
  \includegraphics[width=0.98\columnwidth]{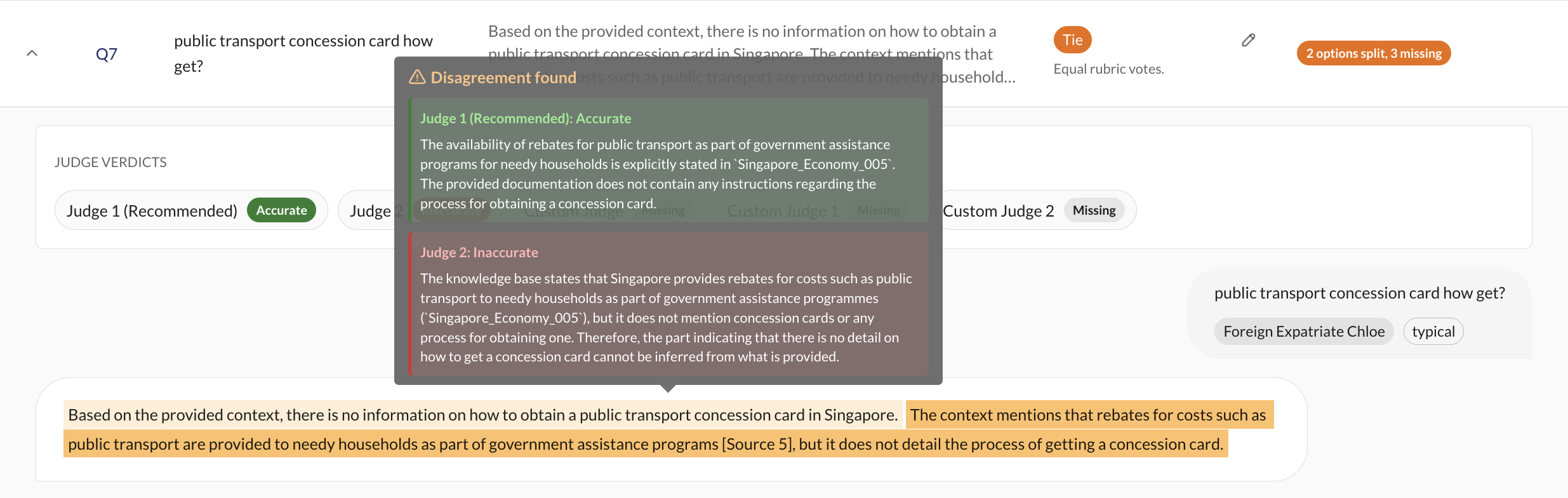}
  \caption{Judge disagreements for error analysis.}
  \label{fig:disagreement}
\end{figure}

This jury design reduces reliance on a single automated evaluator, allows smaller or lower-cost models to be used when they pass the local alignment threshold, and exposes judge disagreements. These disagreements are useful for error analysis because they can indicate ambiguous examples or responses requiring closer human review (Figure~\ref{fig:disagreement}).

\section{Testing with Business Users}
\label{sec:testing-business-users}

A three-week pilot was conducted across four organizational use cases to assess whether \textsc{Kaleidoscope}'s workflow fit practical evaluation work and to collect design feedback. The pilot examined perceived usability, workflow fit, and interpretation of evaluation artifacts; it was not a controlled comparison against manual evaluation or existing tools, nor a validation of evaluation correctness. Table~\ref{tab:pilot-use-cases} summarizes the use cases; Appendix~\ref{app:pilot-setup} provides data-collection details.

\begin{table}[h]
    \centering
    \small
        \begin{tabular}{@{}p{0.32\columnwidth}p{0.62\columnwidth}@{}}
        \toprule
        \textbf{Use case} & \textbf{Evaluation focus} \\
        \midrule
        Finance claims bot & Accuracy for finance queries. \\
        HR bot & Realistic and relevant questions. \\
        Procurement bot & Contract grounding and accuracy. \\
        Staff assistant & App usefulness and tone. \\
        \bottomrule
        \end{tabular}
    \caption{Pilot use cases.}
    \label{tab:pilot-use-cases}
\end{table}

The pilot involved eight users/testers (one to three per application) from four teams across the four use cases. Six users completed the post-pilot questionnaire (75\% response rate). Each tester was asked to generate an evaluation set of at least 30 test cases and run it across three snapshots of the target. Throughout the pilot, participants completed approximately 12 evaluation runs, 180 generated test cases, and 40 human-reviewed outputs under the default 20\% review setting. 

\subsection{User Feedback}

Questionnaire responses and follow-up interviews suggested that participants perceived \textsc{Kaleidoscope} as reducing the manual effort required to organize and inspect evaluations, particularly through automated test orchestration and the human-review interface. In the post-pilot questionnaire, 83\% of respondents (5 of 6) reported that \textsc{Kaleidoscope} helped them evaluate their applications more efficiently, and all respondents reported considering the judge reliability score when interpreting automated results. These responses are early usability evidence that the workflow artifacts and calibration signal were salient to decision-making.

\subsubsection{Lessons for System Design}
The same qualitative feedback identified practical friction points and mismatches between the initial workflow assumptions and users' evaluation needs, leading to several design revisions.

\paragraph{Revising test-case allocation for realistic coverage.}
The initial generator spread test cases evenly across typical and edge-case queries and in- and out-of-knowledge-base scopes. Pilot users found that this over-represented inputs unlike expected application usage, so we revised the allocation controls and added ``input style'' as a generation dimension to better capture colloquial queries. Ratios are detailed in Appendix~\ref{app:test-generation}.

\paragraph{Grounding target setup with additional context.}
The workflow depends on sufficient application context: users are asked to describe the application and upload relevant knowledge-base content, but pilot users sometimes lacked access to this material or supplied only brief descriptions. To reduce sparse-context failure modes, we added a backend web-search grounding step that supplements setup with public information about the application's domain and organizational goals where available. This fit the pilot setting, where most applications had relevant public information online, but narrower or more specialized use cases still require users to inject application- or domain-specific context.

\paragraph{Guiding users towards multi-judge evaluation.}
Although \textsc{Kaleidoscope} supports comparing several LLM-based judges, most pilot users stayed with the default automated judge rather than configuring additional judges. We attribute this partly to limited familiarity with judge configuration and multi-judge evaluation and therefore retain the jury approach, while improving interface cues to explain when judge disagreements are informative.

\paragraph{Supporting flexible data uploads and exports.}
Pilot users often needed only parts of the end-to-end workflow: some brought existing evaluation datasets for automated scoring, while others wanted generated test cases for offline annotation or analysis. We therefore added import and export points so users can upload test cases and export generated evaluations. This lets \textsc{Kaleidoscope} integrate with existing evaluation processes rather than require teams to restructure around the tool. 

\section{Deployment Considerations}
We outline deployment considerations when \textsc{Kaleidoscope} moves toward an integrated testing product.

\paragraph{Operational Integration.}
Evaluation cost is a practical constraint, especially as AI evaluation is itself becoming a compute bottleneck \citep{evalevalCostsBottleneck}. This concern applies to \textsc{Kaleidoscope}, where cost scales with the number of test cases, scoring granularity, and judge calls. Claim-level scoring and multi-judge evaluation are especially expensive, as they require more model calls per evaluation run. Although the workflow can use smaller models and reliability-filtered judging to control token cost, deployed systems should expose and monitor cost controls so teams can trade-off evaluation depth, latency, and expense.

Evaluation quality also depends on how well test generation and scoring are grounded in the application knowledge. The pilot showed that users may not know which context to provide in \textsc{Kaleidoscope}. Therefore, future deployments should support enhanced integration options with applications and their retrieval systems so that evaluations can access relevant context with less manual setup. 

\paragraph{Validation and Calibration at Scale.}
The pilot provided early practical feedback but is not fully representative of the workflow's long-term validity. Use cases were limited, and users were of varying technical backgrounds, most of whom were new to evaluations (Section~\ref{sec:testing-business-users}). Production deployment should test the workflow with a larger user base and higher-risk technical use cases.

A key calibration challenge is balancing review set size against evaluation frequency. If evaluations are run infrequently, reviewers may be willing to annotate a larger set which produces more robust reliability estimates. On the other hand, if evaluations are run frequently, such as after incremental application changes, it becomes impractical for users to review large sets -- and may lead to reviewers skipping the calibration step altogether, relying solely on LLM judge scores. This is useful as a stopgap, but increases risk substantially. 

The current workflow also relies on a limited annotation process. A stronger setup would involve multiple annotators and aggregation of their ratings, enabling the system to capture annotator variation and reduce individual reviewer bias. However, this increases overall manual effort and may slow down testing and governance processes. 

\textsc{Kaleidoscope} therefore treats human calibration and reliability scores as a quality check rather than a golden substitute for automated scoring, and there is the need for applications to configure calibration settings based on their risk level and reviewer tolerance and capacity.

\paragraph{Interpreting Results for Governance.}
\textsc{Kaleidoscope} can inform governance, assurance, or quality-assurance workflows, but its results should be interpreted locally with respect to the user's rubric, test set, and labels rather than as standalone scores. The reliability gate in Section~\ref{subsec:llm-jury} provides a quality check and passes selected judges as scalable scorers, but does not provide guarantees of correctness or coverage. Teams are encouraged to select reliability gate thresholds that fit their applications' purpose. The same goes for evaluation scores, which depend largely on user-defined acceptable thresholds. There is no single ``best'' threshold for deciding whether an application passes governance requirements. \textsc{Kaleidoscope} can report that an application scored a certain percentage on a given rubric, but whether that score is acceptable largely depends on the product and its risk level. For instance, the same aggregate "Accuracy" score may be acceptable for a low-risk criterion but insufficient for a high-risk one. Governance teams should hence interpret scores alongside the rubrics, test cases, human annotations, judge failures, and domain-specific risk requirements.

\paragraph{Expanding Evaluation Scope.}
Finally, the current workflow is strongest for input-output evaluation: it can evaluate any application where inputs can be sent to the target system and collect outputs for review. This covers many practical AI applications, but does not fully capture the intermediate behaviors in more complex systems. Agentic applications may involve tool calls, multi-agent interactions, long-horizon planning; while RAG applications may require checks on retrieval accuracy. Evaluating such systems holistically would require richer traces for more comprehensive diagnostics. The team is actively working on extending the workflow in these directions.

\section{Conclusion and Future Extensions}
\textsc{Kaleidoscope} shows that contextual evaluation for real-world AI applications requires more than fixed benchmarks or automated judge scores. Teams need support for constructing representative tests, defining application-specific rubrics, reviewing outputs efficiently, and calibrating automated judges. We present \textsc{Kaleidoscope} as a workflow for contextual, functional evaluation. Our pilot and experiments suggest that persona-based test generation, claim-level review, and judge calibration and analysis can make this process more practical, while keeping the HITL for defining task contexts and staying accountable for automated scores. 

The next phase is to productize these lessons in a broader testing platform, which we are actively working towards. Beyond scaling the workflow, a key extension is to close the loop from diagnosis to improvement. Evaluation results could help teams propose targeted changes to prompts, retrieval settings, application behavior, or judge configurations. Future work should also extend the testing scope beyond input-output evaluation to more complex AI applications with agentic components, multi-turn workflows, or retrieval systems.

\bibliography{custom}

@article{liu2023geval,
  title={G-Eval: NLG Evaluation using GPT-4 with Better Human Alignment},
  author={Liu, Yang and Iter, Dan and Xu, Yichong and Wang, Shuohang and Xu, Ruochen and Zhu, Chenguang},
  journal={arXiv preprint arXiv:2303.16634},
  year={2023},
  doi={10.48550/arXiv.2303.16634}
}

@inproceedings{gupta2022answerability,
  title={Answerability: A custom metric for evaluating chatbot performance},
  author={Gupta, Prakhar and Rajasekar, Adithya A and Patel, Ameet and Kulkarni, Manav and Sunell, Anirudh and Kim, Khyathi and Ganapathy, Karthik and Trivedi, Ashish},
  booktitle={Proceedings of the Second Workshop on Natural Language Generation, Evaluation, and Metrics},
  pages={316--325},
  year={2022},
  publisher={Association for Computational Linguistics},
  doi={10.18653/v1/2022.gem-1.27}
}

@article{adiwardana2020meena,
  title={Towards a Human-like Open-Domain Chatbot},
  author={Adiwardana, Daniel and Luong, Minh-Thang and So, David R and Hall, Jamie and Fiedel, Noah and Thoppilan, Romal and Yang, Zi and Kulshreshtha, Apoorv and Nemade, Gaurav and Lu, Yifeng and Le, Quoc V},
  journal={arXiv preprint arXiv:2001.09977},
  year={2020},
  doi={10.48550/arXiv.2001.09977}
}

@inproceedings{chen2024emotionqueen,
  title={EmotionQueen: A Benchmark for Evaluating Empathy of Large Language Models},
  author={Chen, Yirong and Yan, Shijie and Liu, Sijia and Li, Yinxiao and Xiao, Yanghua},
  booktitle={Findings of the Association for Computational Linguistics: ACL 2024},
  pages={2149--2176},
  year={2024},
  publisher={Association for Computational Linguistics},
  doi={10.18653/v1/2024.findings-acl.128}
}

@misc{liang2023holisticevaluationlanguagemodels,
      title={Holistic Evaluation of Language Models}, 
      author={Percy Liang and Rishi Bommasani and Tony Lee and Dimitris Tsipras and Dilara Soylu and Michihiro Yasunaga and Yian Zhang and Deepak Narayanan and Yuhuai Wu and Ananya Kumar and Benjamin Newman and Binhang Yuan and Bobby Yan and Ce Zhang and Christian Cosgrove and Christopher D. Manning and Christopher Ré and Diana Acosta-Navas and Drew A. Hudson and Eric Zelikman and Esin Durmus and Faisal Ladhak and Frieda Rong and Hongyu Ren and Huaxiu Yao and Jue Wang and Keshav Santhanam and Laurel Orr and Lucia Zheng and Mert Yuksekgonul and Mirac Suzgun and Nathan Kim and Neel Guha and Niladri Chatterji and Omar Khattab and Peter Henderson and Qian Huang and Ryan Chi and Sang Michael Xie and Shibani Santurkar and Surya Ganguli and Tatsunori Hashimoto and Thomas Icard and Tianyi Zhang and Vishrav Chaudhary and William Wang and Xuechen Li and Yifan Mai and Yuhui Zhang and Yuta Koreeda},
      year={2023},
      eprint={2211.09110},
      archivePrefix={arXiv},
      primaryClass={cs.CL},
      url={https://arxiv.org/abs/2211.09110}, 
}

@article{krishna2023longeval,
  title={LongEval: Guidelines for Human Evaluation of Faithfulness in Long-form Summarization},
  author={Krishna, Kalpesh and Bransom, Emily and Kuehl, Bailey and Iyyer, Mohit and Dasigi, Pradeep and Cohan, Arman and Lo, Kyle},
  journal={arXiv preprint arXiv:2301.13298},
  year={2023},
  doi={10.48550/arXiv.2301.13298}
}

@misc{openai2025evals,
  title={How evals drive the next chapter in AI for businesses},
  author={{OpenAI}},
  year={2025},
  howpublished={\url{https://openai.com/index/evals-drive-next-chapter-of-ai/}},
  note={Accessed 2026-06-02}
}

@article{pignatiello2020decisionFatigue,
  title={Decision fatigue: A conceptual analysis},
  author={Pignatiello, Grant A. and Martin, Richard J. and Hickman, Ronald L. Jr.},
  journal={Journal of Health Psychology},
  volume={25},
  number={1},
  pages={123--135},
  year={2020},
  doi={10.1177/1359105318763510}
}

@article{karim2025annotationAgents,
  title={Transforming Data Annotation with AI Agents: A Review of Architectures, Reasoning, Applications, and Impact},
  author={Karim, Md Monjurul and Khan, Sangeen and Van, Dong Hoang and Liu, Xinyue and Wang, Chunhui and Qu, Qiang},
  journal={Future Internet},
  volume={17},
  number={8},
  pages={353},
  year={2025},
  doi={10.3390/fi17080353}
}

@misc{gu2025surveyJudge,
  title={A Survey on LLM-as-a-Judge},
  author={Gu, Jiawei and Jiang, Xuhui and Shi, Zhichao and Tan, Hexiang and Zhai, Xuehao and Xu, Chengjin and Li, Wei and Shen, Yinghan and Ma, Shengjie and Liu, Honghao and Wang, Saizhuo and Zhang, Kun and Lin, Zhouchi and Zhang, Bowen and Ni, Lionel and Gao, Wen and Wang, Yuanzhuo and Guo, Jian},
  year={2025},
  eprint={2411.15594},
  archivePrefix={arXiv},
  primaryClass={cs.CL}
}

@misc{ai2contextualized2025,
  title={Contextualized Evaluations: Judging language model responses to underspecified queries},
  author={Malaviya, Chaitanya and Chang, Joseph Chee and Roth, Dan and Iyyer, Mohit and Yatskar, Mark and Lo, Kyle},
  year={2025},
  howpublished={\url{https://allenai.org/blog/contextualized-evaluations}},
  note={Accessed 2026-06-02}
}

@misc{openaiEvalsGithub,
  title={Evals: A framework for evaluating LLMs and LLM systems},
  author={{OpenAI}},
  year={2026},
  howpublished={\url{https://github.com/openai/evals}},
  note={Accessed 2026-06-02}
}

@misc{inspectAI,
  title={Inspect: An open-source framework for large language model evaluations},
  author={{UK AI Security Institute and Meridian Labs}},
  year={2026},
  howpublished={\url{https://inspect.aisi.org.uk/}},
  note={Accessed 2026-06-02}
}

@misc{nvidiaNemotronPersonas,
  title={Nemotron-Personas Collection: Region-Specific Synthetic Persona Datasets},
  author={{NVIDIA}},
  year={2026},
  howpublished={\url{https://huggingface.co/collections/nvidia/nemotron-personas}},
  note={Accessed 2026-06-03}
}

@misc{promptfoo,
  title={Promptfoo: Build secure AI applications},
  author={{Promptfoo}},
  year={2026},
  howpublished={\url{https://www.promptfoo.dev/}},
  note={Accessed 2026-06-02}
}

@misc{ibmEvalAssist,
  title={EvalAssist},
  author={{IBM}},
  year={2026},
  howpublished={\url{https://ibm.github.io/eval-assist/}},
  note={Accessed 2026-06-02}
}

@misc{arizeJury,
  title={LLM-as-a-Jury: What It Is and How To Implement},
  author={{Arize AI}},
  year={2026},
  howpublished={\url{https://arize.com/llm-as-a-jury/}},
  note={Accessed 2026-06-02}
}

@misc{evalevalCostsBottleneck,
  title={AI evals are becoming the new compute bottleneck},
  author={Ghosh, Avijit and Mai, Yifan and Channing, Georgia and Choshen, Leshem},
  year={2026},
  howpublished={\url{https://huggingface.co/blog/evaleval/eval-costs-bottleneck}},
  note={Accessed 2026-06-03}
}

@article{jain2025beyond,
  title={Beyond Consensus: Mitigating the Agreeableness Bias in {LLM} Judge Evaluations},
  author={Jain, Suryaansh and Ahmed, Umair Z. and Sahai, Shubham and Leong, Ben},
  journal={arXiv preprint arXiv:2510.11822},
  year={2025}
}

@article{haldar2025rating,
  title={Rating Roulette: Self-Inconsistency in {LLM}-As-A-Judge Frameworks},
  author={Haldar, Rajarshi and Hockenmaier, Julia},
  journal={arXiv preprint arXiv:2510.27106},
  year={2025}
}

@misc{saligrama2026groundTruthProcess,
  title={Ground truth is a process, not a dataset},
  author={Saligrama, Venkatesh},
  year={2026},
  howpublished={Amazon Science, \url{https://www.amazon.science/blog/ground-truth-is-a-process-not-a-dataset}},
  note={Accessed 2026-07-02}
}

\appendix

\section{Key Features of the Workflow}
\label{app:interface-screenshots}

The key pages in the \textsc{Kaleidoscope} workflow consist of target setup, rubric definition, persona configuration, and scoring.

\subsection{Target Setup}
\label{app:target-setup}
\begin{figure}[h]
  \centering
  \includegraphics[width=0.98\columnwidth]{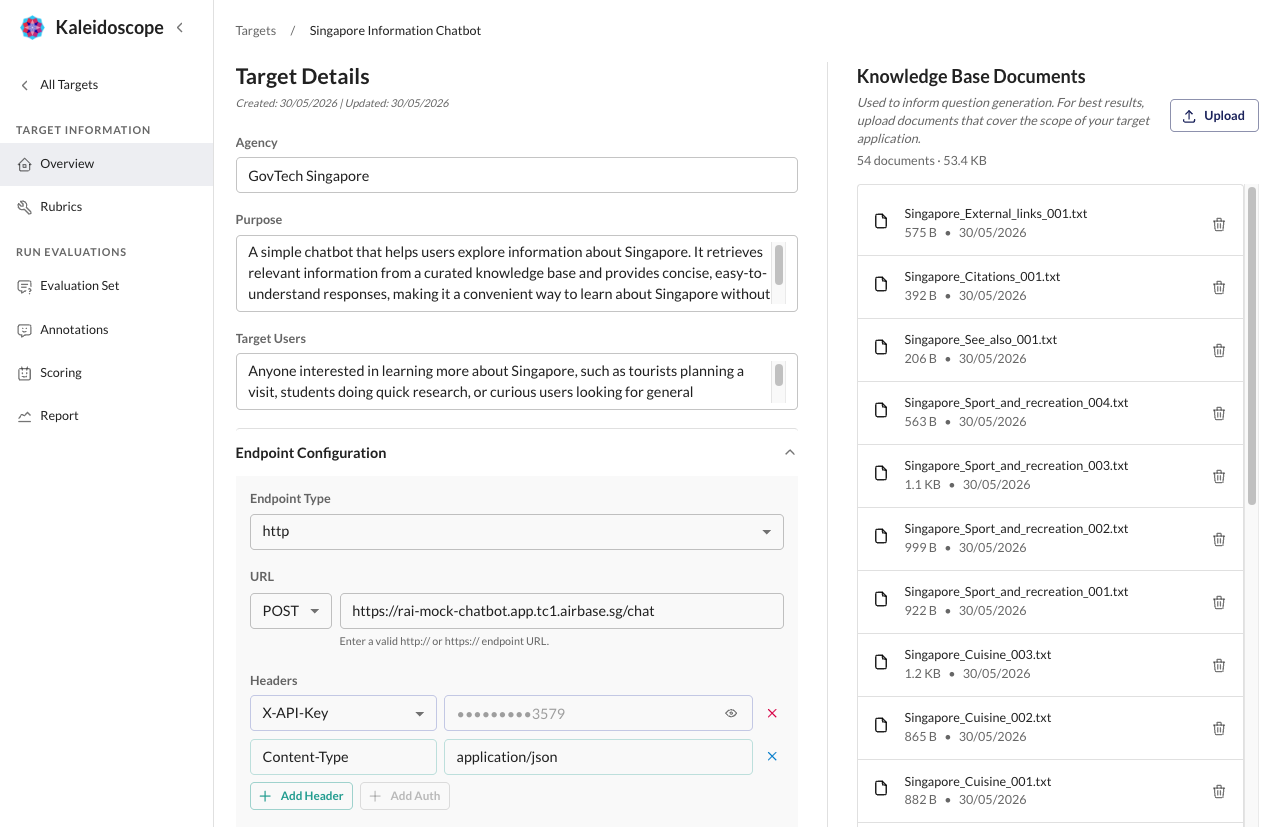}
  \caption{Target setup page, where users provide application context for test generation and evaluation.}
  \label{fig:target-setup-page}
\end{figure}

Figure~\ref{fig:target-setup-page} showcases the fields collected by \textsc{Kaleidoscope} during target setup.

\subsection{Defining Rubrics}
\label{app:defining-rubrics}
\begin{figure}[h]
  \centering
  \includegraphics[width=0.98\columnwidth]{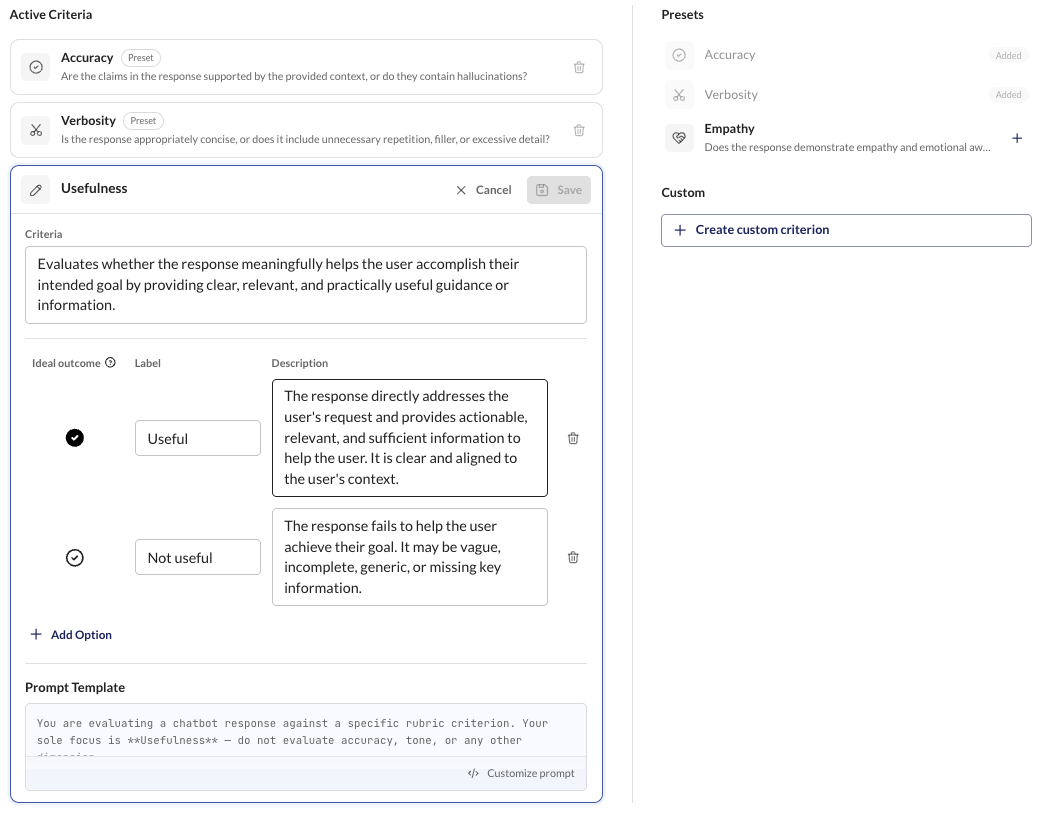}
  \caption{Rubric configuration page, where users define preset or custom evaluation criteria.}
  \label{fig:rubrics-page}
\end{figure}

Users can select from preset rubrics or create custom rubrics in Figure~\ref{fig:rubrics-page}.

\subsection{Persona Configuration and Test Generation}
\label{app:test-generation}
\begin{figure}[h]
  \centering
  \begin{minipage}{0.48\columnwidth}
    \centering
    \includegraphics[width=\linewidth]{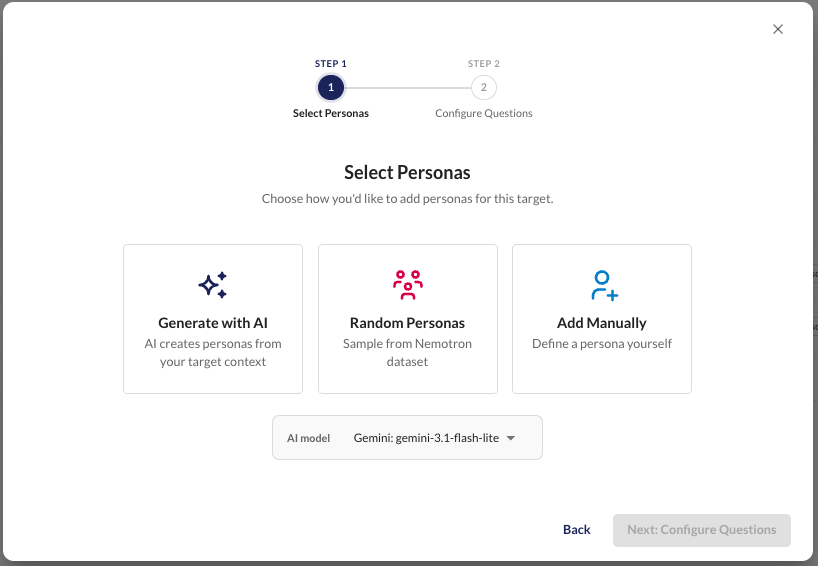}
  \end{minipage}\hfill
  \begin{minipage}{0.48\columnwidth}
    \centering
    \includegraphics[width=\linewidth]{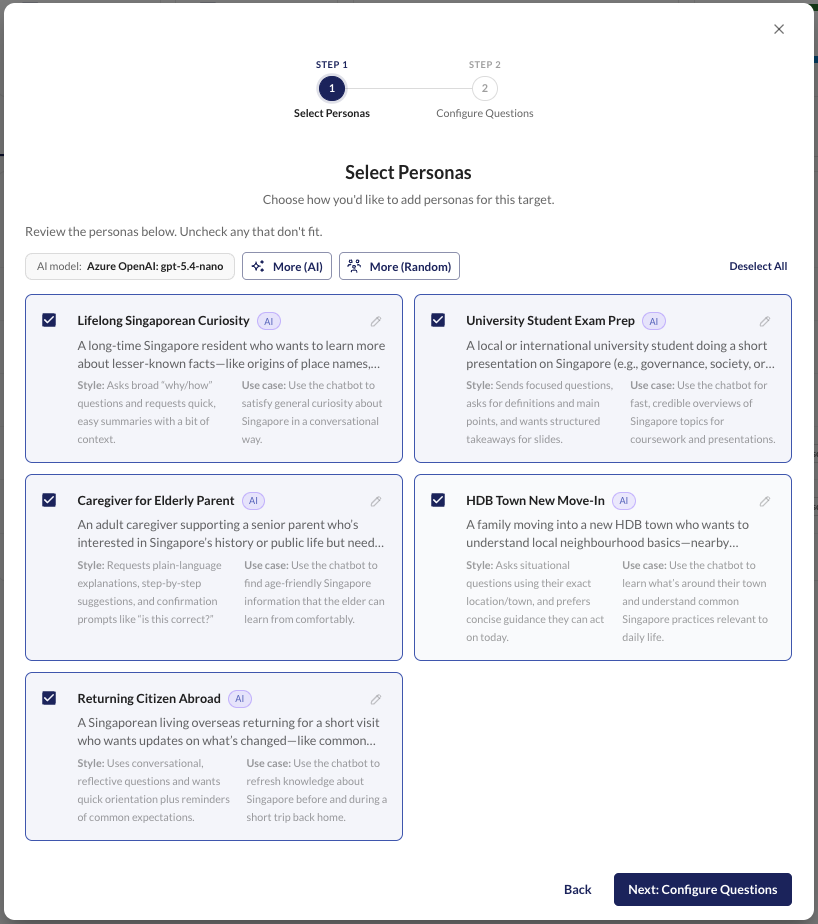}
  \end{minipage}
  \caption{Persona configurations.}
  \label{fig:persona-pages}
\end{figure}

Users have three options to curate personas in Figure~\ref{fig:persona-pages}. 

Table~\ref{tab:generation-allocation} shows the default test allocation, tuned after pilot feedback to better reflect real-world usage.

\begin{table}[h]
\centering
\small
\begin{tabular}{@{}llr@{}}
\toprule
\textbf{KB setting} & \textbf{Type and scope} & \textbf{Ratio} \\
\midrule
With KB & Typical + in-KB & 70\% \\
        & Edge + in-KB & 15\% \\
        & Typical + out-of-KB & 10\% \\
        & Edge + out-of-KB & 5\% \\
\midrule
Without KB & Typical + out-of-KB & 80\% \\
           & Edge + out-of-KB & 20\% \\
\bottomrule
\end{tabular}
\caption{Default allocation for generated questions.}
\label{tab:generation-allocation}
\end{table}

\subsection{Automated Scoring}
\label{app:automated-scoring}
\begin{figure}[h]
  \centering
  \includegraphics[width=0.98\columnwidth]{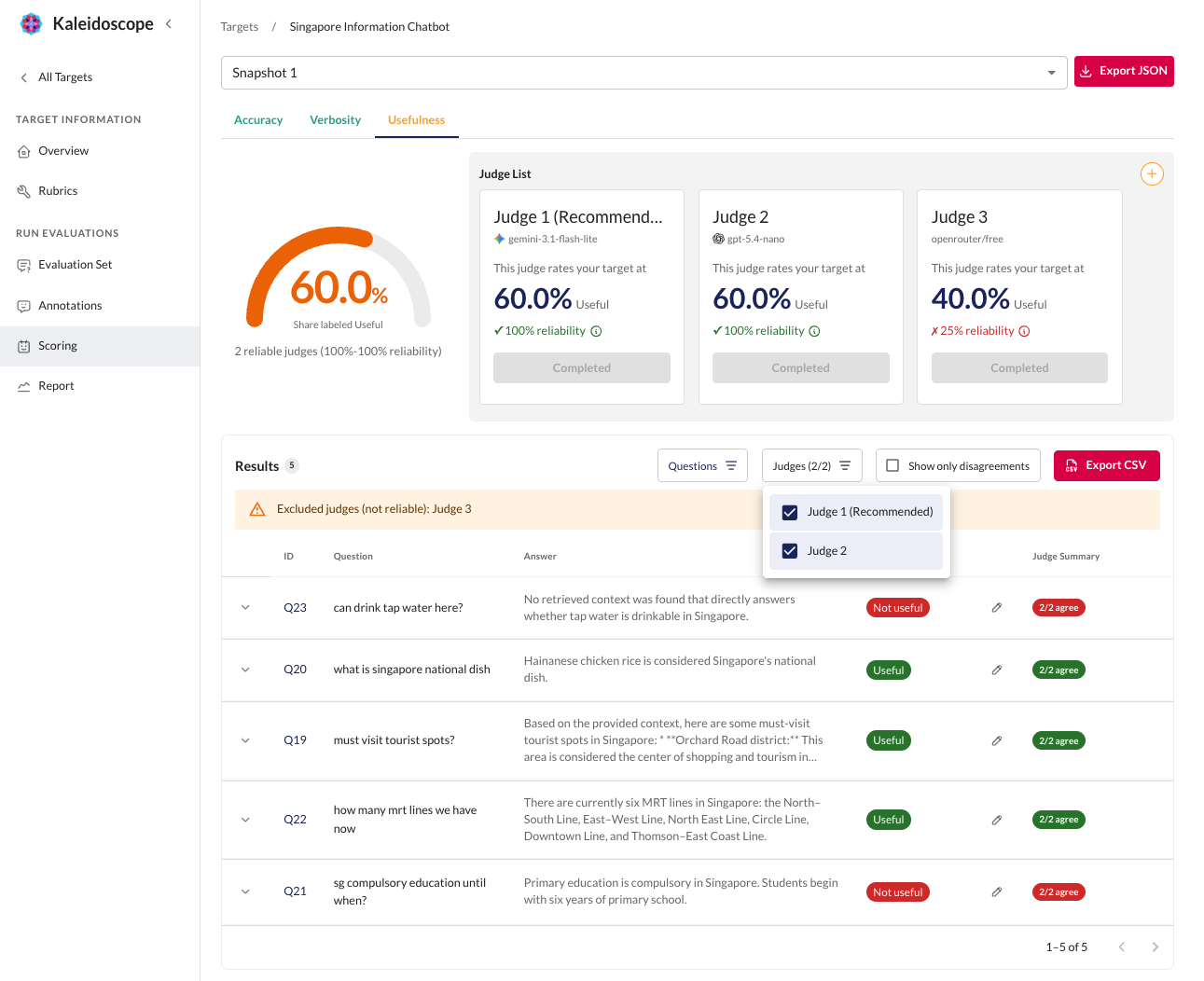}
  \caption{Scoring page, where users inspect automated scores and evaluation results.}
  \label{fig:scoring-page}
\end{figure}

Figure~\ref{fig:scoring-page} displays the aggregated judge result and error analysis table. Users are able to filter for specific personas, test case dimensions, judges, and view disagreements. 

\subsubsection{Claim-Level Scoring}
\label{app:claim-level-scoring}

\begin{figure}[h]
  \centering
  \includegraphics[width=0.8\columnwidth]{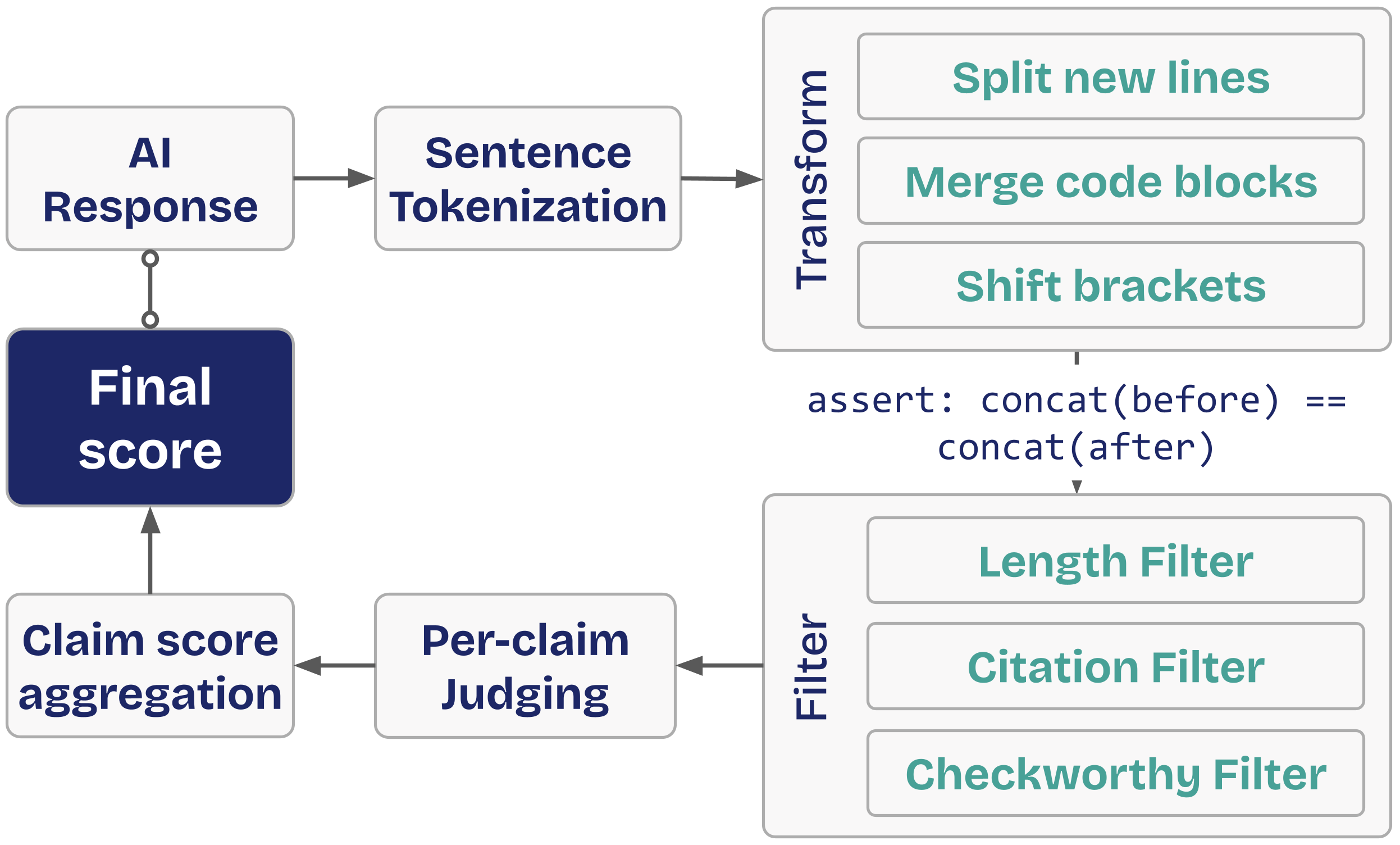}
  \caption{Claim-level scoring pipeline.}
  \label{fig:checkworthy-pipeline}
\end{figure}

For accuracy-oriented metrics, \textsc{Kaleidoscope} can score a response at the claim level rather than holistically judging the full chatbot response. Figure~\ref{fig:checkworthy-pipeline} summarises this pipeline.

\paragraph{Sentence tokenization.}
Responses are chunked with NLTK's sentence tokenization. This step is intentionally deterministic and preserves context.

\paragraph{Transform and post-processing.}
Tokenised sentences are post-processed into meaningful chunks.

\paragraph{Filtering and checkworthiness.}
This stage removes claims that should bypass the accuracy judge, and consists of (i) simple deterministic filters that handle short fragments and citation artifacts, and (ii) an LLM-based checkworthiness step, which decides whether a candidate claim is worth checking. This reduces claim checks on headings, filler text, or irrelevant phrases. In addition, the checkworthy model (user-configured) sees the candidate claim in local context rather than in isolation.

\paragraph{Judging and aggregation.}
Checkworthy claims are then judged against the knowledge base or retrieval citations, producing a binary accuracy verdict for each claim. Aggregation is strict: any unsupported claim makes the response inaccurate.

\section{Rubric and Judge Experiments}
\label{app:rubric-experiments}

In the process of designing how rubrics and judges are set up, we ran experiments across different architectures. These experiments were conducted iteratively during system development. Their purpose was to compare candidate rubric and judge architectures, diagnose sources of misalignment, and inform the design of Kaleidoscope. They should therefore be interpreted as formative design experiments on a shared development dataset rather than as a comprehensive benchmark of the evaluated models.

\subsection{Calibration Data}
\label{app:calibration-dataset}


The calibration set consists of 108 question--answer pairs from four AI systems
in RubricBench,\footnote{\url{https://huggingface.co/datasets/govtech/RubricBench}}
annotated by two human annotators across 14 dimensions.

\textbf{Questions} were synthetically generated using \textsc{Kaleidoscope} and \textbf{Answers} were produced by 4 AI systems, each with two prompt styles.

\begin{table}[h]
    \centering
    \footnotesize
    \setlength{\tabcolsep}{3pt}
    \renewcommand{\arraystretch}{0.92}
    \begin{tabularx}{\linewidth}{@{}l>{\raggedright\arraybackslash}X>{\raggedright\arraybackslash}X@{}}
        \toprule
        & \textbf{Simple} & \textbf{Complex} \\
        \midrule
        
        \textbf{Factual}
        &
        \textbf{Driving Tutor Bot} (\textit{N=30}): \newline
        factual and serious tutor $\rightarrow$
        strict ``no-nonsense'' instructor with sarcasm
        &
        \textbf{Housing Bot} (\textit{N=30}): \newline
        neutral policy specialist $\rightarrow$
        emotional, outspoken, prejudiced officer
        \\
        
        \textbf{Subjective}
        &
        \textbf{Career Friend Bot} (\textit{N=28}): \newline
        helpful career support staff $\rightarrow$
        formal, exhaustive career executive
        &
        \textbf{Responsible AI Bot} (\textit{N=20}): \newline
        encouraging AI consultant $\rightarrow$
        strict, authoritative mentor
        \\
        
        \bottomrule
    \end{tabularx}
    \caption{AI systems in the calibration set. Each cell shows the baseline style  $\rightarrow$ contrastive style.}
    \label{tab:calibration-systems}
\end{table}

The two annotators independently evaluated all 108 question--answer pairs across the 14 dimensions. We did not combine their annotations into a single consensus label. Instead, each judge was evaluated separately against each annotator, and the resulting accuracy, positive-class binary F1, precision, recall, and Cohen's $\kappa$ values were averaged across the two comparisons. 

\subsection{Candidate Judges and Scoring}
\label{app:judge-models}

Table~\ref{tab:judge-models} lists the eight candidate judge models tested across the experiment phases. The judge outputs were scored against human annotations using accuracy, positive-class binary F1, precision, recall, and Cohen's $\kappa$. Each judge configuration was run three times, and judge outputs were combined by majority vote before calculation of alignment metrics.

\begin{table}[h]
    \centering
    \small
    \begin{tabular}{@{}ll@{}}
        \toprule
        \textbf{Model} & \textbf{Provider} \\
        \midrule
        GPT-4.1 & OpenAI \\
        GPT-5-mini & OpenAI \\
        Claude Haiku 4.5 & Anthropic \\
        Grok 4.1 Fast & xAI \\
        Gemini 2.5 Flash & Google \\
        Gemini 3 Flash & Google \\
        Gemini 3.1 Flash Lite & Google \\
        GLM-5 & Zhipu AI \\
        \bottomrule
    \end{tabular}
    \caption{Candidate judge models evaluated across the rubric experiments.}
    \label{tab:judge-models}
\end{table}

\paragraph{Metric hierarchy for judge selection.} When selecting the best judge per metric, we prioritized: (1) Cohen's $\kappa$, as it corrects for chance agreement and penalizes degenerate classifiers; (2) F1, as a secondary check on problem-detection ability; (3) recall, as a tiebreaker when kappa and F1 are comparable --- preferring judges that catch more problems over those that are conservative.

We used Cohen’s $\kappa$ for comparative analysis but the Kaleidoscope workflow uses Macro $F1$ for the runtime reliability gate because it is easier to interpret across user-defined categorical rubrics and gives equal weight to each class.

\subsection{Phase 1: Multi-Metric Bracket Approach}
\label{app:rubric-phase1}

\textbf{Objective.} In initial experiments, we explored a multi-metric judge architecture intended to reduce evaluation cost. Evaluation metrics were grouped into three families or "brackets" of evaluation criteria, with one judge call producing structured labels for every metric in the corresponding bracket. 

\begin{itemize}[leftmargin=*]
    \item \textbf{Relevancy:} sensibleness, specificity, context recall, and instruction following.
    \item \textbf{Voice:} bias, coherence, fluency, empathy, clarity, professionalism, structure, and verbosity.
    \item \textbf{Accuracy:} faithfulness and factual correctness.
\end{itemize}

This reduced the number of judge calls required for evaluation. 

\textbf{Hypothesis.} Related evaluation metrics can be assessed together within a single bracket-level judge call without materially reducing evaluation quality.

\paragraph{Setup.} The Relevancy and Voice brackets were both tested across 8 models. The Accuracy bracket was excluded from the experiments as it requires a different evaluation pipeline described in Section~\ref{app:claim-level-scoring}

For each metric, judge alignment was calculated separately against each annotator and averaged across the two annotators. The resulting per-metric scores were then averaged equally to obtain the overall bracket-level Cohen's $\kappa$ and Macro $F1$.

\begin{table}[h]
    \centering
    \small
    \begin{tabular}{@{}lccc@{}}
        \toprule
        \textbf{Bracket} & \textbf{Best $\kappa$} & \textbf{Best F1} & \textbf{Best Judge} \\
        \midrule
        Relevancy & 0.239 & 0.596 & GLM-5 \\
        Voice & 0.376 & 0.675 & Claude Haiku 4.5 \\
        \bottomrule
    \end{tabular}
    \caption{Best judge per bracket (Phase 1, multi-metric).}
    \label{tab:rubric-phase1-brackets}
\end{table}

\begin{table}[h]
    \centering
    \small
    \begin{tabular}{@{}lccc@{}}
        \toprule
        \textbf{Metric} & \textbf{Best $\kappa$} & \textbf{Best F1} & \textbf{Best Judge} \\
        \midrule
        Bias & 0.818 & 0.842 & GPT-4.1 \\
        Clarity & 0.648 & 0.777 & Claude Haiku 4.5 \\
        Professionalism & 0.540 & 0.660 & Claude Haiku 4.5 \\
        Structure & 0.514 & 0.652 & Gemini 2.5 \\
        Empathy & 0.485 & 0.775 & GPT-4.1 \\
        Verbosity & 0.337 & 0.805 & Claude Haiku 4.5 \\
        Fluency & 0.220 & 0.346 & Gemini 2.5 \\
        Coherence & 0.174 & 0.276 & Claude Haiku 4.5 \\
        \bottomrule
    \end{tabular}
    \caption{Per-metric results within the Voice bracket (Phase 1). Metrics are ordered by kappa.}
    \label{tab:rubric-phase1-voice-permetric}
\end{table}

Overall agreement was low for Relevancy and moderate for Voice. However, the aggregated scores hid substantial differences between the individual metrics evaluated in the same judge call. After examining performance separately for each metric in the Voice bracket, we found that performance varied substantially across metrics. Agreement was relatively strong for bias and clarity but weaker for verbosity, fluency, and coherence. The best-performing model differed across metrics, suggesting that no single judge model was consistently strongest across the dimensions in the Voice bracket.

These findings suggested that low per-metric performance may reflect both the multi-metric architecture and the inherent ambiguity or class distribution of individual metrics. Phase 1 alone could not distinguish between these explanations, which led us to investigate whether prompt iteration could improve performance.

\subsection{Phase 2: Prompt Iteration on Brackets}
\label{app:rubric-phase2}

\textbf{Objective.} Phase~2 tested whether the multi-metric bracket architecture could be retained while improving performance through prompt optimization. 

\textbf{Hypothesis.} Adding calibration notes, stricter definitions, and reasoning structures would improve bracket performance.

Across three rounds, we retained the same multi-metric voice and relevancy brackets while progressively revising their prompts. Round~1 used baseline prompts with concise metric definitions and label criteria. Round~2 expanded these definitions with concrete failure cases, calibration guidance against defaulting to positive labels, dimension-specific checks, and evidence-based justifications. Round~3 built on variants of the Round~2 prompts, testing explicit verification steps, clearer decision thresholds, illustrative examples, and a shorter, compressed formulation. 

For each round and bracket, Table~\ref{tab:rubric-phase2} reports the judge--prompt configuration with the highest bracket-level Cohen's $\kappa$ among the configurations evaluated in that round. The Round~3 value therefore represents the best-performing model--prompt variant, rather than an average across prompt variants.

\begin{table}[h]
    \centering
    \small
    \resizebox{\columnwidth}{!}{%
    \begin{tabular}{@{}lcccccc@{}}
        \toprule
        \textbf{Bracket} & \textbf{R1 $\kappa$} & \textbf{R2 $\kappa$} & \textbf{R3 $\kappa$} & \textbf{$\Delta\kappa$} & \textbf{R1 F1} & \textbf{R3 F1} \\
        \midrule
        Relevancy & 0.239 & 0.205 & 0.192 & $-$19.5\% & 0.596 & 0.567 \\
        Voice & 0.376 & 0.344 & 0.373 & $-$0.9\% & 0.675 & 0.674 \\
        \bottomrule
    \end{tabular}%
    }
    \caption{Prompt iteration across three rounds (Phase 2). Performance degraded or plateaued.}
    \label{tab:rubric-phase2}
\end{table}

Prompt iteration degraded Relevancy and failed to improve Voice. One possible explanation is \textit{scope overload}. Each bracket-level judge must maintain distinct decision boundaries for several metrics within a single call. However, rubric ambiguity, label distribution, and prompt-specific effects may also contribute.

\subsection{Phase 3: Single-Metric Judges}
\label{app:rubric-phase3}

\textbf{Objective.} Phase~3 tested whether narrowing the judge's scope to one metric would improve alignment with human annotations.

\textbf{Hypothesis.} Dedicated metric-level judges would produce better alignment by allowing full focus on a single dimension.

\paragraph{Metric selection.} Empathy and verbosity were selected based on two criteria: (1) organizational priority - these were the top dimensions flagged as important by pilot adopters; and (2) moderate-to-low multi-metric kappa in Phase~1 (empathy: 0.485, verbosity: 0.337), indicating room for improvement under a different architecture.

\paragraph{Setup.} We developed three single-metric prompt variants for each metric. For each prompt variant, we evaluated multiple candidate models and selected the model with the highest Cohen's $\kappa$. Tables~\ref{tab:rubric-phase3-empathy} and~\ref{tab:rubric-phase3-verbosity} report the best-performing judge configuration for each prompt iteration. 

\begin{table}[h]
\centering
\small
\resizebox{\columnwidth}{!}{%
\begin{tabular}{@{}llccccc@{}}
\toprule
\textbf{Pmt} & \textbf{Best Judge} & \textbf{$\kappa$} & \textbf{F1} & \textbf{Rec.} & \textbf{Prec.} & \textbf{Acc.} \\
\midrule
1 & Gemini 3 Flash & 0.559 & 0.824 & 0.767 & 0.892 & 0.787 \\
2 & GPT-5-mini & 0.494 & 0.744 & 0.602 & 0.977 & 0.732 \\
3 & GPT-5-mini & 0.545 & 0.794 & 0.688 & 0.941 & 0.769 \\
\bottomrule
\end{tabular}%
}
\caption{Single-metric judge results for \textit{empathy} (\texttt{pos\_label} = ``not empathetic'').}
\label{tab:rubric-phase3-empathy}
\end{table}

\begin{table}[h]
\centering
\small
\resizebox{\columnwidth}{!}{%
\begin{tabular}{@{}llccccc@{}}
\toprule
\textbf{Pmt} & \textbf{Best Judge} & \textbf{$\kappa$} & \textbf{F1} & \textbf{Rec.} & \textbf{Prec.} & \textbf{Acc.} \\
\midrule
1 & Gemini 3.1 Flash Lite & 0.376 & 0.810 & 0.878 & 0.756 & 0.736 \\
2 & Gemini 3 Flash & 0.332 & 0.604 & 0.443 & 0.953 & 0.625 \\
3 & Gemini 3 Flash & 0.340 & 0.795 & 0.851 & 0.750 & 0.718 \\
\bottomrule
\end{tabular}%
}
\caption{Single-metric judge results for \textit{verbosity} (\texttt{pos\_label} = ``verbose'').}
\label{tab:rubric-phase3-verbosity}
\end{table}

\paragraph{Comparison with the multi-metric baseline.}
For each metric, we selected the prompt--model configuration with the highest Cohen's $\kappa$ across the three variants, using F1 as a secondary measure of classification performance. This resulted in Prompt~1 with Gemini 3 Flash for empathy ($\kappa=0.559$, F1$=0.824$) and Prompt~1 with Gemini 3.1 Flash Lite for verbosity ($\kappa=0.376$, F1$=0.810$).

We then compared these configurations with the corresponding Phase~1 results, where empathy and verbosity were evaluated alongside the other dimensions within the multi-metric Voice judge. Both comparisons use the same metrics, evaluation examples, and human annotations. However, they compare the strongest prompt--model configuration found under each architecture, and the selected judge model may differ between the multi-metric and single-metric setups. The observed differences therefore reflect the combined effects of prompt scope and model selection and should not be attributed solely to single-metric prompting.

\begin{table}[h]
    \centering
    \small
    \resizebox{\columnwidth}{!}{%
    \begin{tabular}{@{}lcccccc@{}}
        \toprule
        \textbf{Metric} & \textbf{Multi. $\kappa$} & \textbf{Sing. $\kappa$} & \textbf{$\Delta\kappa$} & \textbf{Multi. F1} & \textbf{Sing. F1} & \textbf{$\Delta$F1} \\
        \midrule
        Empathy & 0.485 & 0.559 & +15.2\% & 0.775 & 0.824 & +6.3\% \\
        Verbosity & 0.337 & 0.376 & +11.6\% & 0.805 & 0.810 & +0.6\% \\
        \bottomrule
    \end{tabular}%
    }
    \caption{Best single-metric judges compared with the same metrics evaluated within the multi-metric Voice judge.}
    \label{tab:rubric-phase3-comparison}
\end{table}

The stronger empathy result is consistent with the hypothesis that narrowing the judging scope can help apply metric-specific criteria more consistently, although the comparison also reflects differences in the selected judge model. The verbosity results were more modest. The multi-metric judge had already achieved a high F1 score, and the dedicated judge primarily improved agreement rather than overall classification performance.

Overall, the results provide partial support for the hypothesis. Single-metric configurations achieved higher alignment with the human annotations, but the gains were modest and the resulting $\kappa$ values remained moderate. This suggests that judging scope may contribute to misalignment, but is unlikely to be its only source.

\subsection{Phase 4: LLM-Based Prompt Augmentation}
\label{sec:phase4}

\textbf{Objective.} Phases~1--2 (\ref{app:rubric-phase2}) showed that general-purpose multi-metric templates produced inconsistent performance across criteria, while Phase~3 (\ref{app:rubric-phase3}) showed that dedicated single-metric prompts could improve alignment for selected metrics. However, manually developing a dedicated prompt for every possible user-defined criterion is not scalable.

We therefore explored an LLM-based prompt augmentation setup. Given a user-defined rubric, an augmentation model expands the rubric into a metric-specific judge prompt containing additional definitions and calibration guidance. A fixed judge model then evaluates using the augmented prompt.

\begin{figure*}[t]
  \centering
  \includegraphics[width=\textwidth]
  {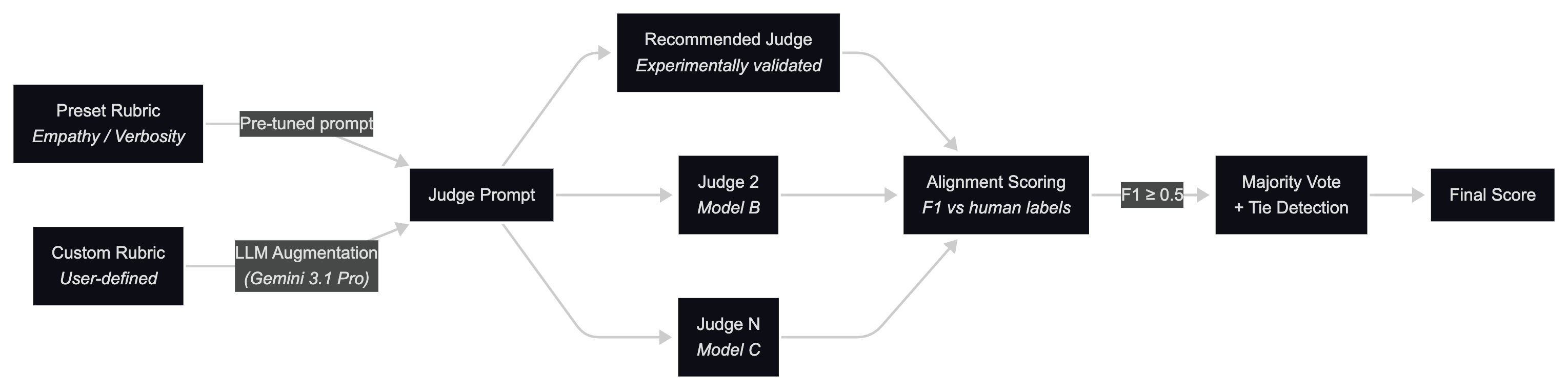}
  \caption{Hybrid rubric and judge workflow in \textsc{Kaleidoscope}.
  Preset rubrics use pre-tuned prompts, while custom rubrics are expanded through LLM-based augmentation. Candidate judges are evaluated against human annotations, and only judges that meet the configured reliability threshold contribute to majority-vote aggregation.}
  \label{fig:hybrid-judge-design}
\end{figure*}

Figure~\ref{fig:hybrid-judge-design} illustrates how this hybrid design is incorporated into the broader judging workflow.

\textbf{Hypothesis.} An augmentation model can generate a
metric-specific judge prompt that improves alignment relative to
evaluating the same criterion within a general-purpose multi-metric
prompt.

\paragraph{Metric selection.} Structure and specificity were selected because: (1) both were organizational priorities flagged during pilot discussions, and (2) their properties are relatively orthogonal, providing a stress test of the augmentation approach across metric types.

\paragraph{Setup.} We evaluated four augmentation models---GPT-5.4,
Claude Sonnet 4.6, Gemini 3.1 Pro, and GLM-5---across three levels of
rubric detail:

All three levels represented the same underlying metric, label set, and intended decision boundary. The detailed and few-shot versions added guidance without intentionally changing what the rubric measured.

\begin{itemize}[leftmargin=*]
    \item \textbf{Barebones:} a brief description of the metric and its
    labels;
    \item \textbf{Detailed:} the same underlying metric expanded with
    decision criteria, clarifications, and edge cases; and
    \item \textbf{Few-shot:} the detailed definition supplemented with
    annotated examples illustrating the intended application of the
    labels.
\end{itemize}

Gemini 3.1 Flash Lite was held fixed as the downstream judge due to its performance in Phase~3 (~\ref{app:rubric-phase3}), to isolate the effects of the augmentation model. We evaluated two metrics: structure (\texttt{pos\_label} = ``poorly structured'') and specificity (\texttt{pos\_label} = ``generic''). This produced 24 configurations: four augmentation models $\times$ three rubric-detail levels $\times$ two metrics.

Tables~\ref{tab:phase4-structure} and
\ref{tab:phase4-specificity} report the five highest-$\kappa$
configurations for each metric.

\begin{table}[h]
    \centering
    \small
    \begin{tabular}{@{}lccc@{}}
        \toprule
        \textbf{Augmenter + Complexity} & \textbf{$\kappa$} & \textbf{F1} & \textbf{Acc.} \\
        \midrule
        Gemini 3.1 Pro + Detailed & 0.548 & 0.692 & 0.801 \\
        GPT-5.4 + Few-shot & 0.514 & 0.652 & 0.792 \\
        Gemini 3.1 Pro + Few-shot & 0.488 & 0.626 & 0.782 \\
        GLM-5 + Detailed & 0.482 & 0.613 & 0.782 \\
        Gemini 3.1 Pro + Barebones & 0.481 & 0.640 & 0.773 \\
        \bottomrule
    \end{tabular}
    \caption{Top 5 augmentation results for \textit{structure}.}
    \label{tab:phase4-structure}
\end{table}
    
\begin{table}[h]
    \centering
    \small
    \begin{tabular}{@{}lccc@{}}
        \toprule
        \textbf{Augmenter + Complexity} & \textbf{$\kappa$} & \textbf{F1} & \textbf{Acc.} \\
        \midrule
        GPT-5.4 + Barebones & 0.336 & 0.446 & 0.769 \\
        Gemini 3.1 Pro + Barebones & 0.329 & 0.473 & 0.750 \\
        Gemini 3.1 Pro + Few-shot & 0.298 & 0.465 & 0.732 \\
        Claude Sonnet + Detailed & 0.293 & 0.471 & 0.722 \\
        GLM-5 + Detailed & 0.289 & 0.427 & 0.741 \\
        \bottomrule
    \end{tabular}
    \caption{Top 5 augmentation results for \textit{specificity}.}
    \label{tab:phase4-specificity}
\end{table}

The strongest augmentation configuration differed across the two
metrics. For structure, Gemini 3.1 Pro with the detailed rubric
achieved the highest agreement ($\kappa=0.548$). For specificity,
GPT-5.4 with the barebones rubric performed best
($\kappa=0.336$). More detailed rubric inputs therefore did not
consistently produce stronger augmented prompts.

\paragraph{Comparison with the multi-metric baseline.}
For each metric, we selected the augmented configuration with the
highest Cohen's $\kappa$ and compared it with the metric's Phase~1
performance within the multi-metric judge.

\begin{table}[h]
    \centering
    \small
    \resizebox{\columnwidth}{!}{%
    \begin{tabular}{@{}l|ccc|ccc@{}}
        \toprule
        \textbf{Metric} & \textbf{Multi. $\kappa$} & \textbf{Aug. $\kappa$} & \textbf{$\Delta\kappa$} & \textbf{Multi. F1} & \textbf{Aug. F1} & \textbf{$\Delta$F1} \\
        \midrule
        Structure & 0.514 & 0.548 & +6.6\% & 0.652 & 0.692 & +6.2\% \\
        Specificity & 0.366 & 0.336 & $-$8.1\% & 0.484 & 0.446 & $-$7.7\% \\
        \bottomrule
    \end{tabular}%
    }
    \caption{Best augmented-prompt configuration compared with the same
metric evaluated within the Phase~1 multi-metric judge.}
    \label{tab:phase4-comparison}
\end{table}

Prompt augmentation produced mixed results. For structure, the best augmented prompt increased $\kappa$ from 0.514 to 0.548 and F1 from 0.652 to 0.692, modestly outperforming the multi-metric baseline. 

For specificity, however, $\kappa$ decreased slightly from 0.366 to 0.336 and $F1$ decreased from 0.484 to 0.446. The weaker specificity result should also be interpreted alongside the lower inter-annotator agreement for this metric ($\kappa$=0.317, compared with 0.666 for structure). Because judges were evaluated separately against each annotator and the resulting metrics were averaged, inconsistent human decision boundaries make it more difficult for a judge to achieve strong average agreement across both annotators. The specificity result may therefore reflect ambiguity in the rubric or its application, in addition to limitations of the augmented prompt.

The results provide partial support for the hypothesis. LLM-based
augmentation can produce competitive metric-specific prompts, but its
effectiveness depends on the criterion, augmentation model, and amount
of rubric detail. No single augmentation configuration performed best
across both metrics.

\paragraph{Design implication.}
These findings suggest that augmentation should not replace validated metric-specific presets. Instead, they support a hybrid design. Kaleidoscope provides preset rubrics for commonly used metrics that have been tested and refined across user applications. For custom user-defined rubrics, prompt augmentation provides a scalable way to generate an initial metric-specific judge prompt, which users can review and edit.

Both preset and augmented judges are treated as starting points and are not fixed evaluators. Kaleidoscope allows users to measure judge alignment against their own human annotations and iteratively refine the rubric, prompt, model, or judge configuration. In future work, the reviewed examples and alignment results accumulated through this process could support a more automated feedback loop, in which application-specific data is used to propose or optimize judge prompts before they are re-evaluated against human labels.

\section{Pilot Setup Details}
\label{app:pilot-setup}

\begin{figure}[!t]
  \centering
  \includegraphics[width=0.8\columnwidth]{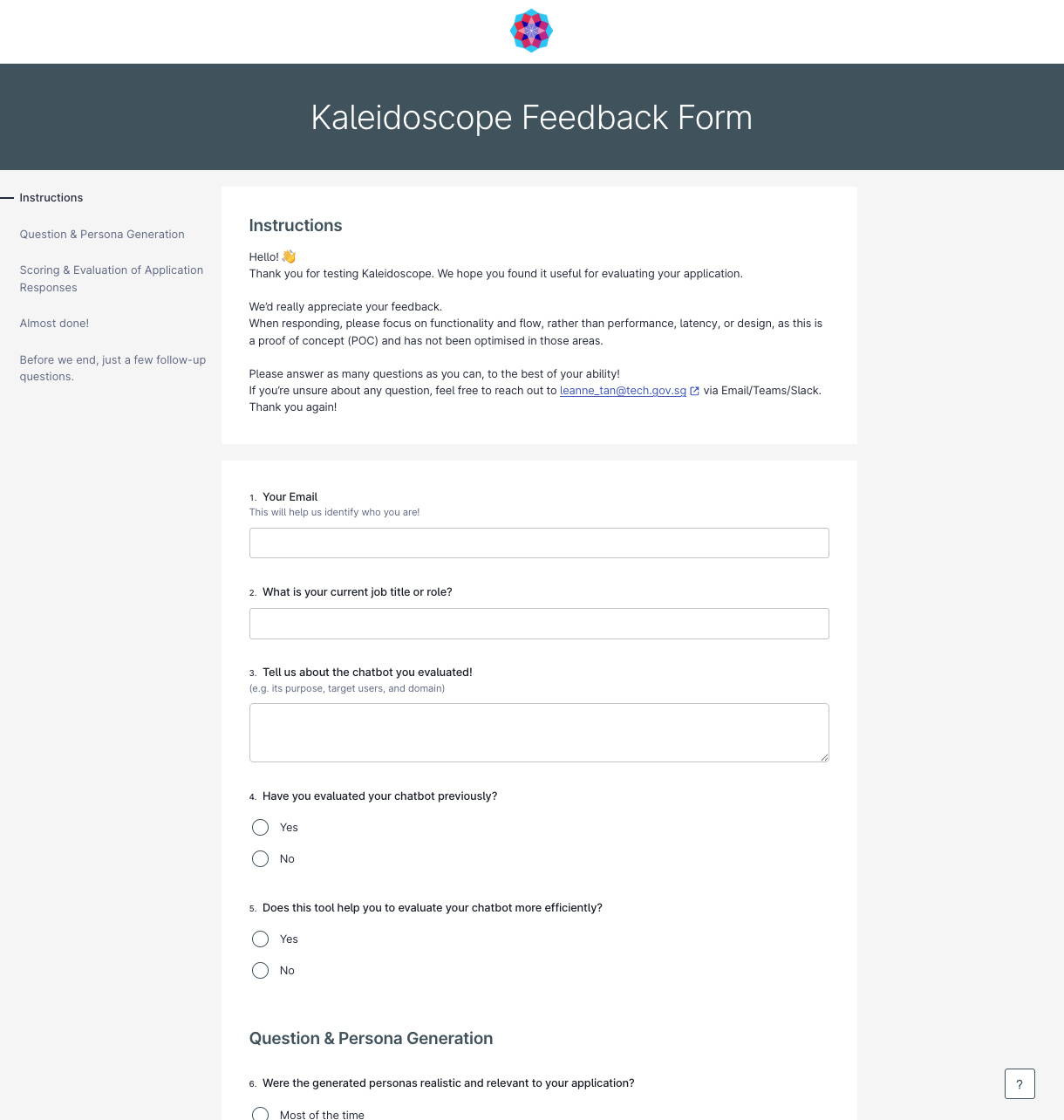}

  \vspace{0.3em}
  \includegraphics[width=0.8\columnwidth]{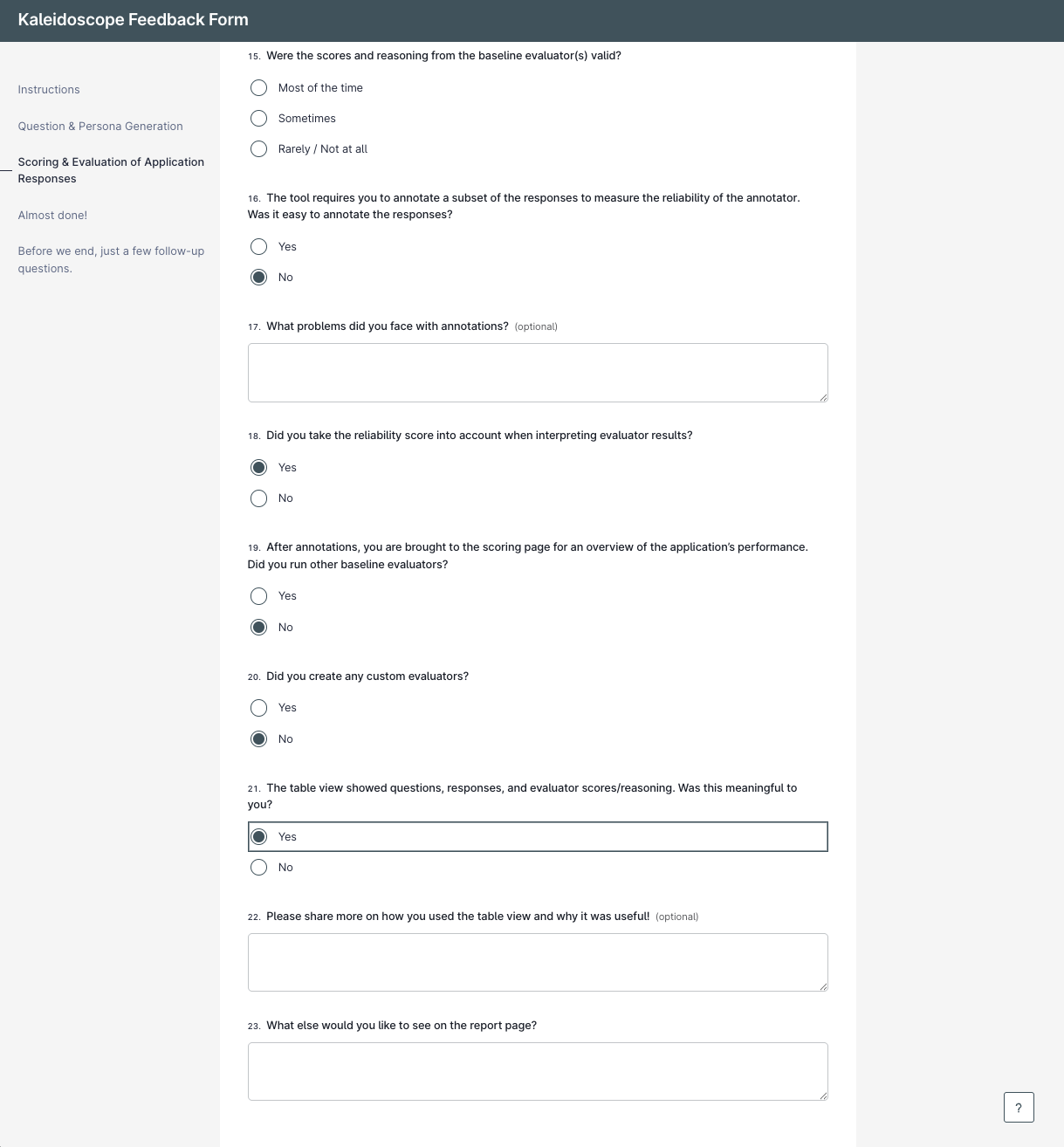}
  \caption{Post-pilot feedback form used to collect ratings and qualitative comments from pilot users.}
  \label{fig:pilot-feedback-form}
\end{figure}

Feedback was collected through a post-pilot questionnaire and follow-up interviews. Beyond numeric ratings, we collected detailed comments from users and interviewed them about their usage patterns. Due to privacy considerations, details of the individual applications, test cases, and evaluation results are redacted. Figure~\ref{fig:pilot-feedback-form} shows the feedback form used for post-pilot collection.

The pilot provided useful early feedback on the \textsc{Kaleidoscope} workflow, but was limited in scale and duration. It also captured users' first-time interaction with the system, where some friction may have been due to limited familiarity with AI evaluation concepts, such as rubric configuration and multi-judge evaluation. The next phase will test the workflow as part of a broader testing product, allowing \textsc{Kaleidoscope} to be evaluated in a more operational setting with broader use cases.

\end{document}